\documentclass{Interspeech}
\usepackage{cite}
\usepackage{amsmath,amssymb,amsfonts}
\usepackage{algorithmic}
\usepackage{graphicx}
\usepackage{textcomp}
\usepackage{xcolor}

\usepackage{multirow}
\usepackage{hyperref}
\usepackage{booktabs}
\usepackage{array}
\usepackage{lineno}
\usepackage{tablefootnote}
\usepackage{tikz}
\usepackage{pgfplots}
\usepackage{threeparttable}

\interspeechcameraready

\title{Language-Aware Prompt Tuning for Parameter-Efficient Seamless Language Expansion in Multilingual ASR
\thanks{*Corresponding author: Hao Huang}
\thanks{This work was supported by National Natural Science Foundation of China (62466055) and the National Key R\&D Program of China
(2020AAA0107902).}
}

\author[affiliation={1}]{Hongli}{Yang}
\author[affiliation={2}]{Sheng}{Li}
\author[affiliation={1,3,4,*}]{Hao}{Huang}
\author[affiliation={1}]{Ayiduosi}{Tuohan}
\author[affiliation={5}]{Yizhou}{Peng}

\affiliation{School of Computer Science and Technology}{ Xinjiang University}{China}
\affiliation{}{Institute of Science Tokyo}{Japan}
\affiliation{}{Joint International Research Laboratory of Silk Road Multilingual Cognitive Computing}{China}
\affiliation{}{Xinjiang Key Laboratory of Multi-lingual Information Technology}{China}
\affiliation{College of Computing and Data Science}{Nanyang Technological University}{Singapore}

\email{hongli@stu.xju.edu.cn, huanghao@xju.edu.cn}


\usepackage{comment}

\begin{document}

\maketitle

\begin{abstract}

Recent advancements in multilingual automatic speech recognition (ASR) have been driven by large-scale end-to-end models like Whisper. However, challenges such as language interference and expanding to unseen languages (language expansion) without degrading performance persist. This paper addresses these with three contributions: 1) Entire Soft Prompt Tuning (Entire SPT), which applies soft prompts to both the encoder and decoder, enhancing feature extraction and decoding; 2) Language-Aware Prompt Tuning (LAPT), which leverages cross-lingual similarities to encode shared and language-specific features using lightweight prompt matrices; 3) SPT-Whisper, a toolkit that integrates SPT into Whisper and enables efficient continual learning. Experiments across three languages from FLEURS demonstrate that Entire SPT and LAPT outperform Decoder SPT by 5.0\% and 16.0\% in language expansion tasks, respectively, providing an efficient solution for dynamic, multilingual ASR models with minimal computational overhead.

\end{abstract}

\section{Introduction}

Recent years have witnessed significant advancements in Automatic Speech Recognition (ASR) models, especially for high-resource languages, achieving remarkable performance\cite{DBLP:conf/icml/GravesFGS06,DBLP:conf/icassp/ChanJLV16,DBLP:conf/icassp/KimHW17,li2022recent}. As a crucial element of speech interaction in multilingual AI assistants, multilingual speech recognition has attracted considerable research interest, fueling efforts to broaden its applications\cite{zhang2023speechgpt,DBLP:conf/iclr/TangYSC000M024,fang2025llamaomni}.

The development of large-scale multilingual ASR models, such as Whisper\cite{radford2023robust}, Google USM\cite{zhang2023google}, and MMS\cite{pratap2024scaling}, has revolutionized the field. These foundational models enable the creation of customized multilingual speech recognition models tailored to specific languages, unlocking new possibilities for cross-lingual communication and application development\cite{yang2024adapting, talafha23_interspeech,liu2024exploration}.  However, two significant challenges still persist in multilingual ASR: (1) language interference, arising from confusion in all-in-one architectures, and (2) the difficulty of expanding to unseen languages, especially low-resource ones, without degrading the performance of existing ones.

To address language interference, prior works have explored various strategies, such as leveraging language ID information or designing language-specific modules\cite{DBLP:conf/icassp/LiY0DW24,sun2023building,DBLP:conf/interspeech/WangMLD23}, including dedicated encoders for each language. Other approaches employ pruning strategies to create submodels for individual languages\cite{DBLP:conf/icassp/XieLGTSSWJMK24,DBLP:conf/icassp/LuHQWM22} or propose novel sampling methods to mitigate data imbalance. While these methods alleviate language interference to some extent, they often introduce complexities in design and fail to accommodate seamless language expansion.

The integration of new low-resource languages into multilingual ASR models has emerged as a critical research area. Traditional approaches, such as Full Fine-Tuning (FFT), update all parameters of a pre-trained model to adapt to a new task or dataset. While effective, FFT is computationally expensive and prone to catastrophic forgetting (CF) of previously learned languages \cite{jain23_interspeech,yang2023chinese}. To mitigate CF, Continual Learning (CL) methods have been proposed, including prototype-based \cite{michieli2023online}, regularization-based \cite{vander2023rehearsal}, replay-based \cite{cappellazzo2023investigation}, and optimization-based \cite{chaudhryefficient}approaches. In response, Parameter-Efficient Fine-Tuning (PEFT) methods such as Low-Rank Adaptation (LoRA) \cite{hu2021lora,song2024lora,xu2024towards} and Soft Prompt Tuning (SPT) \cite{lester2021power,qian24_interspeech,10447492} focus on adjusting only a subset of model parameters or adding new ones, keeping most of the original parameters fixed. This approach is more computationally efficient than full fine-tuning, enabling language adaptation without compromising performance on previously learned languages. 

Existing approaches like LoRA-Whisper \cite{song2024lora}  enhance multilingual ASR by attaching language-specific low-rank matrices, leveraging cross-lingual similarities to improve new language adaptation. Similarly, Soft Language Code Tuning (SLCT) and Soft Prompt Tuning for the Whisper decoder (Decoder SPT) were proposed as methods to add unseen languages to multilingual ASR models \cite{qian24_interspeech}. SLCT trains only language-specific embeddings, while Decoder SPT uses soft prompts to guide language recognition. However, these methods do not explicitly use modules to store language-specific information, leading to difficulties in achieving effective language discrimination or providing language bias, especially when parameters are shared in a single trainable model. 


\begin{figure*}[tb]
\centering
\includegraphics[width=0.8\linewidth]{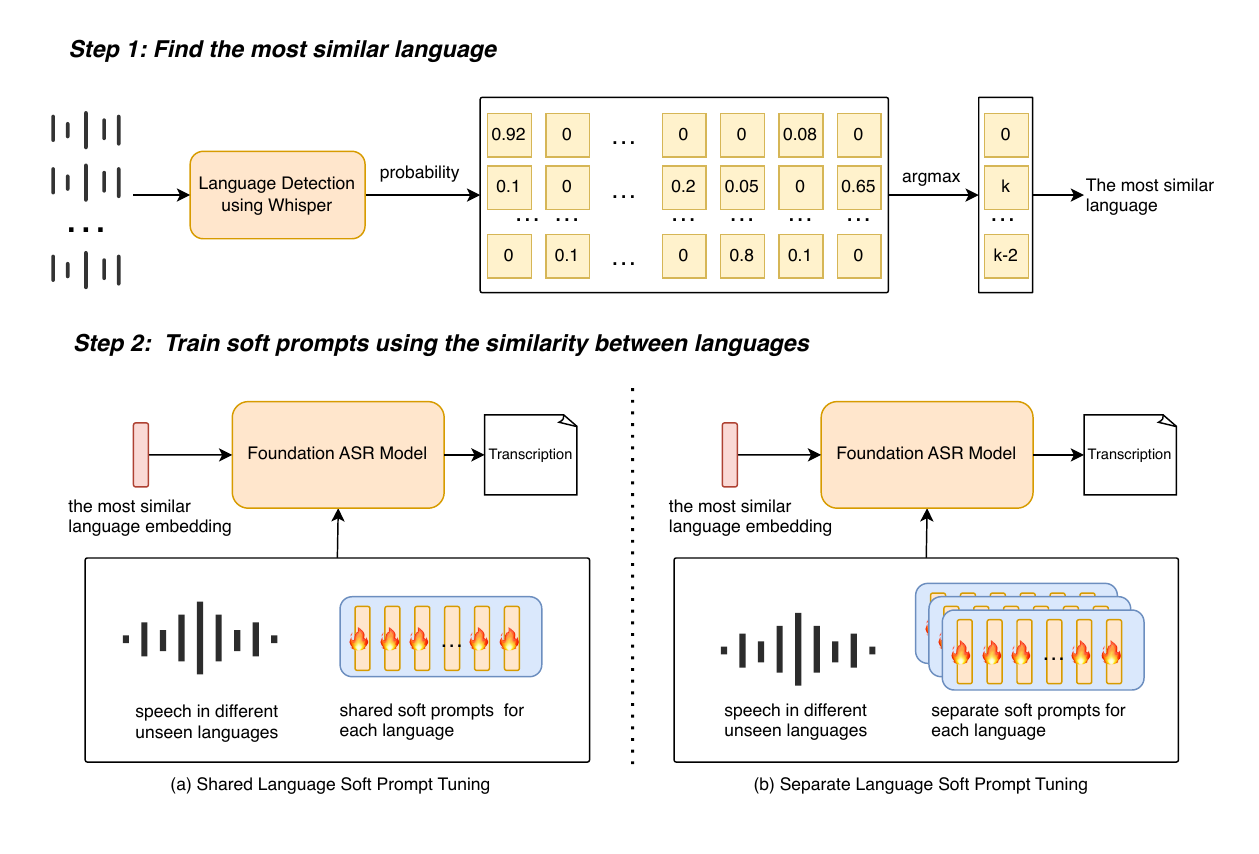}
\caption{Architecture of LAPT in language expansion. Left: Shared language prompt tuning, Right: Separate language prompt tuning.}
\label{fig:LAPT}
\end{figure*}

\begin{figure*}[tb]
\centering
\includegraphics[width=1.0\linewidth]{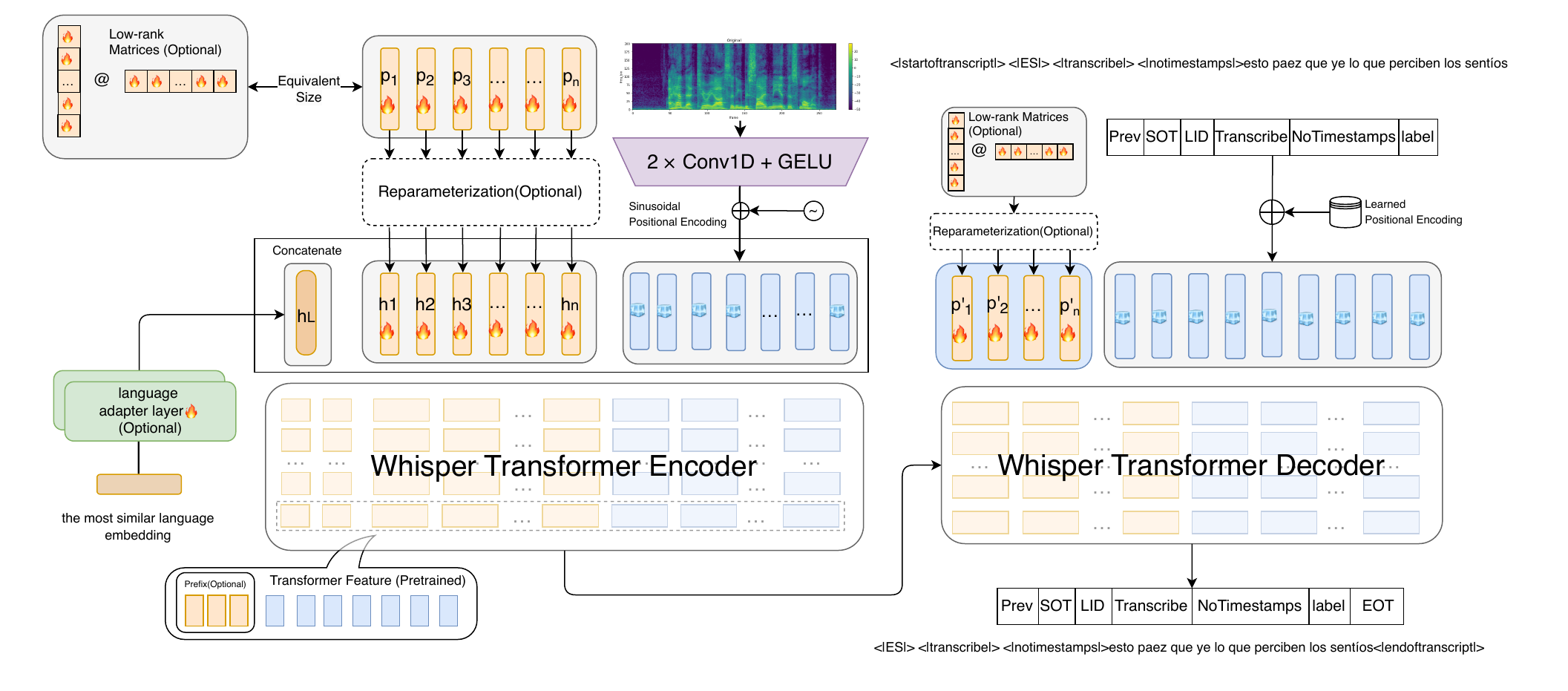}
\caption{Overview of the SPT-Whisper architecture proposed for parameter-efficient continual learning.}
\label{fig:SPT-Whisper}
\end{figure*}

To overcome these limitations, we extend Decoder SPT to the entire model for Whisper (Entire SPT) and propose Language-Aware Prompt Tuning (LAPT), a novel approach that leverages Soft Prompt Tuning and cross-lingual similarities for seamless language expansion. By capitalizing on the linguistic relationships between new and base languages, LAPT stores shared information within the Whisper model while capturing language-specific details in dedicated soft prompt matrices. When incorporating a new language, a unique soft prompt matrix is assigned, ensuring minimal impact on the performance of existing languages. Specifically, we introduce two variants of LAPT: shared language prompt tuning and separate language prompt tuning, enabling flexible and efficient language adaptation.

In addition to the methodological contributions, we develop SPT-Whisper\footnote{\scriptsize \url{https://github.com/paper-submit-code/SPT-Whisper}}, an open-source toolkit that integrates P-Tuning v2\cite{liu2021p}, ResMLP\cite{razdaibiedina2023residual}, LoPT\cite{guo2024lopt}, and advanced SPT techniques into Whisper. SPT-Whisper supports continual learning for multilingual ASR, allowing users to customize models for new languages with minimal computational overhead. Although Whisper serves as the exemplar in this work, the foundational model is not limited to Whisper and can be extended to other open-source speech recognition models.

\section{Methods}

\subsection{Multilingual ASR and Entire SPT}


Applying SPT in multilingual ASR expands language support while mitigating catastrophic forgetting (CF). For seen languages, soft prompts are not needed, as the model’s parameters remain fixed during SPT, allowing standard decoding without degrading performance. For new languages, a unique soft prompt matrix is appended to the Whisper model’s input. The original model retains shared language knowledge, while language-specific nuances are encoded within the corresponding embeddings and soft prompts.

Applying SPT to Whisper involves designing separate soft prompt sequences for both the encoder and the decoder. Given a sequence of acoustic features $X \in \mathbb{R}^{l \times e}$ and a decoder embedding matrix $E \in \mathbb{R}^{|v| \times e}$, we determine the optimal soft prompts $P \in \mathbb{R}^{n \times e}$ for the encoder input and $P^{\prime} \in \mathbb{R}^{n \times e}$ for the decoder input. Here, $l$ is the input speech length, $|v|$ is the vocabulary size, $n$ is the length of the soft prompt sequence, and $e$ is the embedding dimension.

We experimented with different implementations of SPT by applying soft prompts to various components of the Whisper model: specifically, we tested adding soft prompts to only the encoder, only the decoder, or both. For encoder adaptation, the matrix $[P, X] \in \mathbb{R}^{(n+l) \times e}$ is constructed by concatenating the soft prompt matrix $P$ with the input features $X$. This allows the encoder to attend to both the original acoustic features and the learned prompt information. For decoder adaptation, the soft prompts $P^{\prime}$ are inserted into the decoder input at the position of the $\langle\text{Prev}\rangle$ special token. The decoder’s input sequence includes special tokens such as the start-of-sequence, language ID, and task tokens. The resulting modified input is represented as $[P^{\prime}, \text{embed}(g)] \in \mathbb{R}^{(n+4) \times e}$, where $g$ denotes the sequence of special tokens. These inserted soft prompts provide additional contextual guidance for the decoding process.

\subsection{Language-Aware Prompt Tuning}

A common approach to incorporating language information in multilingual speech recognition models is to represent it as a one-hot vector or embedding, which is then concatenated with the acoustic features to form the input. However, in our proposed approach, we aim to leverage the pre-existing language information in a more efficient and adaptable manner. Consequently, we introduce two effective methods for
language expansion, namely Shared Language Prompt Tuning and Separate Language Prompt Tuning as depicted in Fig~\ref{fig:LAPT}. These methods involve two steps.

\textbf{Step 1: Find the most similar language}  
To integrate a new language into an existing model, the first step is to find the language most similar to the new one from a predefined set of base languages. We begin by randomly selecting $\mathbf{M}$ audio segments from the new language’s dataset. These segments are then processed using the Whisper model to perform language identification, generating a probability distribution across all languages. The relevant probabilities are extracted and normalized, yielding the vector $\boldsymbol{p}_i=\left[p_{i1}, p_{i2}, \cdots, p_{in}\right]; i=1, \cdots, M$, where $p_{ij}$ denotes the probability that the $i$-th audio segment belongs to the $j$-th base language.

To determine the most similar language, we calculate the similarity between the new language and each base language using the following metric:
\begin{equation}
\operatorname{sim}_k=\frac{\sum_{i=1}^M \mathbb{I}\left(k=\underset{j}{\arg \max } p_{i j}\right)}{M} \quad \text { for } k=1, \cdots, n
\end{equation}
where $\mathbb{I}$ is the indicator function and ${\operatorname{sim}}_k$ represents the similarity between the new language and the $k$-th base language.

\textbf{Step 2: Continual training on new languages}
Once we have identified the most similar languages, the next step involves training soft prompts using the language similarities. As described in Fig~\ref{fig:LAPT}, building on Whisper’s pre-existing capabilities in language identification and previous research on language token understanding, we attempt to introduce Whisper’s pre-trained language embeddings as language prompts into SPT-based end-to-end multilingual speech recognition. In addition, we use a prompt encoder to encode prompt embedding
with the same dimension as the acoustic features. The prompt
embedding serves as a language prompt, guiding the model
to accurately recognize specific languages with other soft prompts.

In Shared language prompt tuning, shared soft prompts are shared across each language to encourage efficient adaptation and reduce the parameter overhead. In Separate Language Prompt Tuning, separate soft prompts is trained for each new language, providing more granular control over the adaptation process. When the input is a piece of speech of $\mathbf{k}$-th language, it will activate $\mathbf{k}$-th soft prompt and pass through Whisper in the forward pass. Furthermore, language prompts with similarities across languages can be further used to facilitate more effective training for the new languages, thus improving both the accuracy of language identification and the performance of the model on new low-resource languages. 

\subsection{SPT-Whisper}

We implement advanced SPT methods, including LAPT, and provide open-source implementations of all these techniques on SPT-Whisper, as shown in Fig.~\ref{fig:SPT-Whisper}. We aim to provide a fair and practical comparison to advance multilingual ASR.

\section{Experiments}
\label{sec:exp}

\subsection{Dataset and Evaluation Metric}
\label{sec:Dataset}


To ensure the reproducibility of our proposed methods, we conducted experiments on FLEURS benchmark multilingual open-source datasets, which comprise 102 languages, of which 19 languages are not supported by the Whisper model. Each language has about 12 hours of read speech data, with each audio within 30 seconds. Due to limited resources, initial tests were performed on three languages randomly selected from the 19 unsupported ones: Asturian, Sorani Kurdish, and Kabuverdianu. We adopt the Character Error Rates (CER) for evaluating ASR performance. Detailed information of the datasets of these three languages are presented in Table~\ref{tab:fleurs}.

\begin{table}[tb]
\centering
\scriptsize
\caption{FLEURS datasets used in training.}
\label{tab:fleurs}
\setlength{\tabcolsep}{2.60mm} 
\renewcommand\arraystretch{1.0} 
\begin{tabular}{l|c|c|c|c|c|c}
\toprule
\multirow{2}{*}{\textbf{Language}} & \multicolumn{2}{c|}{\textbf{Train}} & \multicolumn{2}{c|}{\textbf{Dev}} & \multicolumn{2}{c}{\textbf{Test}} \\
 & \# & hrs & \# & hrs & \# & hrs \\
\midrule
Asturian & 2511 & 7.53 & 398 & 0.99 & 946 & 2.44 \\
Sorani Kurd & 3,040 & 10.46  & 386  & 1.23  & 922  & 2.99  \\
Kabuverdianu & 2,715 & 10.51  & 366  & 1.32  & 864  & 3.27  \\
\bottomrule
\end{tabular}
\end{table}

\subsection{Implementation Details}
\label{sec:Implementation}

Considering inference speed and limited computational resources, we evaluate our proposed method on not-large foundation models, such as Whisper-small and Whisper-medium. 
FFT employs an initial learning rate of 1e-6, which decays linearly, while LoRA and SPT-based methods use initial rates of 1e-3. LoRA was applied to the attention layers with a rank of 8, and varying rank sizes showed negligible differences in results. We trained the models over 20 epochs with a batch size of 8. The training was conducted on a single NVIDIA RTX3090 GPU with 24GB of VRAM. For decoding, we selected the best model from all epochs based on the dev set and performed a greedy search.


\subsection{Prompt Length and Position Selection}
\label{ssec:length}

The performance of SPT exhibits significant length dependency, as demonstrated by our systematic evaluation on Whisper-small for Asturian recognition.  We evaluated SPT with prompt lengths \{16, 32, 64, 128, 256\}, applied to different components of the model: the encoder (Encoder SPT), decoder (Decoder SPT), and the entire model (Entire SPT). As Table~\ref{tab:prompt_length_position_selection} reveals, two critical patterns emerge: (1) Length Threshold: All configurations achieve optimal performance at 128 tokens. (2) Architecture Sensitivity: Decoder SPT show stronger baseline performance, but Entire SPT integration enables greater error reduction. Therefore, we used a prompt length of 128 and applied Entire SPT in subsequent experiments.


Notably, the best performance is achieved when the soft prompt is applied to the entire model, emphasizing the importance of joint optimization between the encoder and decoder for both acoustic feature extraction and language modeling. This finding is significant because, in traditional ASR models, the encoder typically focuses on extracting speech features from the audio signal, while the decoder uses a language model to convert those features into text. Extending the Decoder SPT to the entire model enables better communication between the two, resulting in improved performance in both tasks.

\begin{table}[tb]
\centering
\scriptsize
\caption{Soft Prompt Tuning Configuration Analysis on Whisper-small: Prompt Length and Position (CER \%)}
\label{tab:prompt_length_position_selection}
\setlength{\tabcolsep}{1.6mm} 
\renewcommand\arraystretch{1.2} 
\begin{tabular}{l|c|c|c}
\toprule
\multirow{1}{*}{\textbf{Prompt Length}} & \multicolumn{1}{c|}{\textbf{Encoder SPT}} & \multicolumn{1}{c|}{\textbf{Decoder SPT}} & \multicolumn{1}{c}{\textbf{Entire SPT}} \\
\midrule
16 Prompt Tokens & 12.76 & 11.61 & 11.77 \\
32 Prompt Tokens & 13.27 & 11.48 & 11.61 \\
64 Prompt Tokens & 13.04 & 11.41 & 10.82 \\
128 Prompt Tokens & 12.33 & 11.91 & 10.31 \\
256 Prompt Tokens & 12.77 & N/A$^{{\spadesuit}}$ & 10.57 \\
\bottomrule
\end{tabular}

\begin{tablenotes}    
    \scriptsize               
    \item  \textit{Note}: ${\spadesuit}$ indicates the decoder does not support 256-token length, due to the decoder's context length limitation. \textit{Entire SPT} with Prompt Length of 256 uses prompts with 256 tokens for the encoder and 128 for the decoder.
\end{tablenotes}

\end{table}

\subsection{Main Results}
\label{ssec:subhead}

\begin{table}[tb]
\centering
\scriptsize
\caption{ASR performance on FLEURS Asturian(ast), Sorani Kurdish(ckb) and Kabuverdianu(kea) test sets with different tuning approaches. (CER \%)}
\label{tab:Comparison}
\setlength{\tabcolsep}{1.60mm}
\renewcommand\arraystretch{1.2}
\begin{tabular}{l|p{2.10cm}|c|c|c|c|c}
\toprule
\multirow{2}{*}{\textbf{ID}} & \multirow{2}{*}{\textbf{Method}} & \multicolumn{4}{c|}{\textbf{Languages}} & \multirow{2}{*}{\textbf{\#Params(M)}} \\
\cline{3-6}
 & & ast & ckb & kea & avg \\
\midrule
B1 & Whisper Small & 14.94  & 57.14  & 36.23 & 36.10 & / \\
B2 & Whisper Medium & 15.01 &  65.36 & 36.20 & 38.86 & / \\
B3 & Whisper Large-v3 & 14.58 & 40.05 & 36.67 & 30.43 & / \\
\midrule
\multicolumn{6}{c}{\textbf{Whisper-small Tuning Results}} \\ 
\midrule
S1 & FFT & 7.14 & 13.16 & 6.32 & 8.87 & 240.58M \\
S2 & LoRA & 10.10 & 18.94 & 9.75 & 12.93 & 1.85M×3  \\
S3 & Shared Entire SPT & 10.12 & 17.37 & 9.81 & 12.43 & 0.17M \\
S4 & \hspace{0.6em}+ LAPT & \textbf{9.99} & 17.35 & 9.68 & 12.34 & 0.96M \\
S5 & Separate Entire SPT & 10.31 & 17.53 & 9.79 & 12.54 & 0.17M×3 \\
S6 & \hspace{0.6em}+ LAPT & 10.01 & \textbf{16.70} & \textbf{9.51} & \textbf{12.07} & 0.96M×3 \\
\midrule
\multicolumn{6}{c}{\textbf{Whisper-medium Tuning Results}} \\ 
\midrule
M1 & FFT & 5.94 & 11.73 & 5.03 & 7.57 & 762.32M \\
M2 & LoRA & 7.52 & 13.76 & 7.06 & 9.45 & 4.94M×3 \\
M3 & Shared Entire SPT & 7.47 & 13.34 & 6.72 & 9.18 & 0.44M \\
M4 & \hspace{0.6em}+ LAPT & 7.45 & 13.25 & 6.70 & 9.13 & 1.49M \\
M5 & Separate Entire SPT & 7.47 & 13.30 & 6.76 & 9.18 & 0.44M×3  \\
M6 & \hspace{0.6em}+ LAPT & \textbf{7.20} & \textbf{13.21} & \textbf{6.67} & \textbf{9.03} & 1.49M×3 \\
\bottomrule
\end{tabular}
\begin{tablenotes}    
    \scriptsize
    \item \textit{Note}: \#Params indicates the number of trainable parameters. “×3” denotes the use of three languages with distinct soft prompt groups.
\end{tablenotes}
\end{table}

Table~\ref{tab:Comparison} presents a comparison of various fine-tuning approaches across three target languages: Asturian (ast), Sorani Kurdish (ckb), and Kabuverdianu (kea), using both Whisper-small and Whisper-medium models. The baseline performance is obtained on different Whisper model versions (B1-B3), while FFT, LoRA, and Entire SPT are fine-tuned on both small and medium models for each language using the respective FLEURS training sets. When applying different tuning methods on Whisper-small, FFT provides the best performance, with an average CER of 8.87\% on Whisper-amall (S1), but at a high computational cost. LoRA (S2) offers a more efficient alternative, yielding a CER of 12.93\% on average, but it still underperforms compared to FFT. Entire SPT provides a balanced approach, with both Shared Entire SPT  (S3) and Separate Entire SPT (S5) educing the average CER to 12.43\% and 12.54\%, respectively, outperforming LoRA. However, the limitation of the Separate SPT approach is that the model size increases as more languages are added, whereas the Shared SPT method addresses this by sharing soft prompts across multiple similar languages.

Integrating LAPT with Entire SPT further improves performance. The combination of Shared Entire SPT and LAPT (S4) results in a minor improvement, reducing the average CER from 12.43\% to 12.34\%. However, the biggest improvement comes from the Separate Entire SPT with LAPT (S6), which reduces the average CER to 12.07\%. This method achieves the lowest CER for Sorani Kurdish (16.70\%), highlighting LAPT’s ability to enhance language-specific recognition.


In Whisper-medium, similar trends are observed, with FFT (M1) delivering the best performance, though at a higher computational cost. The Shared Entire SPT (M3) provides more balanced results compared to the Separate Entire SPT (M5), particularly in terms of parameter efficiency, despite having slightly lower performance. The Separate Entire SPT with LAPT configuration (M6) is expected to outperform LoRA (M2) and Shared Entire SPT with LAPT (M4), making it a more efficient approach for multilingual ASR models.

Overall, experimental results demonstrate that Entire SPT significantly improves the performance of multilingual ASR by applying soft prompts across the entire model. Additionally, the combination of LAPT with Entire SPT provides a scalable solution for expanding ASR models by leveraging language-specific features while minimizing interference during language expansion, all without the computational overhead of FFT.
 
\section{Conclusion}
\label{sec:Conclusion}

We propose a novel approach to address language interference and enable efficient language expansion in multilingual ASR. Entire SPT applies soft prompts to both the encoder and decoder to enhance language recognition, while combining Entire SPT with LAPT leverages cross-lingual similarities for seamless expansion. Additionally, we introduce SPT-Whisper for parameter-efficient continual learning in multilingual ASR, providing a scalable solution for expanding models without compromising performance.

\bibliographystyle{IEEEtran}
\bibliography{mybib}

\end{document}